\documentclass{article}
\usepackage[T1]{fontenc}
\usepackage[preprint]{neurips_2026}
\usepackage{graphicx}
\usepackage{booktabs}
\usepackage{tabularx}
\usepackage{array}
\usepackage{adjustbox}
\usepackage{pdflscape}
\usepackage{hyperref}
\usepackage{url}
\hypersetup{hidelinks}

\makeatletter
\renewcommand{\@noticestring}{}
\renewenvironment{table}
  {\setlength{\abovecaptionskip}{\@neuripsabovecaptionskip}%
   \setlength{\belowcaptionskip}{\@neuripsbelowcaptionskip}%
   \@float{table}}
  {\end@float}
\long\def\@makecaption#1#2{%
  \vskip\abovecaptionskip
  \sbox\@tempboxa{\textbf{#1:} #2}%
  \ifdim \wd\@tempboxa >\hsize
    \textbf{#1:} #2\par
  \else
    \global \@minipagefalse
    \hb@xt@\hsize{\hfil\box\@tempboxa\hfil}%
  \fi
  \vskip\belowcaptionskip}
\makeatother

\title{BenchBench-Cloning:\\Evaluating \mbox{Molecular Cloning} Design\\and Protocol Development}

\author{
Aditya Sivakumar\textsuperscript{\textdagger} \\
Benchling \\
\And
Peter Ciaccia\textsuperscript{*} \\
DNAForge \\
\And
Hassaan Qadir\textsuperscript{*} \\
DNAForge \\
\AND
Ian Hsu\textsuperscript{*} \\
Benchling \\
\And
Ashu Singhal \\
Benchling
}

\newcommand{\passatk}{pass@$k$}
\newcommand{\bestatk}{best-of-$k$}
\newcommand{\figref}[1]{\textbf{Figure~\ref{#1}}}
\newcommand{\figrefs}[2]{\textbf{Figures~\ref{#1} and~\ref{#2}}}
\newcommand{\tabref}[1]{\textbf{Table~\ref{#1}}}
\newcommand{\tabrefs}[2]{\textbf{Tables~\ref{#1} and~\ref{#2}}}
\newcommand{\appref}[1]{\textbf{Appendix~\ref{#1}}}

\newcolumntype{Y}{>{\raggedright\arraybackslash}X}

\begin{document}
\maketitle
\begingroup
\renewcommand{\thefootnote}{\fnsymbol{footnote}}
\footnotetext[1]{Equal contribution.}
\footnotetext[2]{Corresponding author: \texttt{aditya.sivakumar@benchling.com}.}
\endgroup

\begin{abstract}
Molecular cloning is one of the most routine wet-lab workflows, yet existing evaluations of language models in biology mostly test knowledge recall or isolated steps rather than end-to-end experimental planning. We introduce BenchBench-Cloning, a benchmark for large language models comprising 65 molecular cloning tasks grounded in sequence records, laboratory notebooks, and published experimental studies. Planning a molecular cloning experiment requires defining a set of constructs needed for a scientific objective and a construction route to produce them. BenchBench-Cloning contains 50 protocol-development tasks, in which agents produce an executable protocol for a specified construct, and 15 construct-design tasks, in which agents determine the constructs a scientific objective calls for. Protocol-development tasks are evaluated by execution in a cloning simulator; construct-design tasks are evaluated with task-specific deterministic checks and, where needed, rubric-based LLM-as-a-judge evaluation. Across eight systems with repeated attempts per task, GPT-6 Astra scores highest at 30.6\%, with the others between 8.6\% and 22.2\%, and the benchmark remains unsaturated under best-of-ten sampling. As integrating individual cloning decisions into complete, valid plans remains challenging even for frontier models, we expect BenchBench-Cloning to serve as a useful benchmark for tracking progress in experimental planning.
\end{abstract}

\section{Introduction}

Designing and constructing plasmids is a common task in molecular biology. A scientist must determine the required sequence, identify the requisite starting materials, choose a construction strategy, and specify the reactions needed to produce the desired product. The choice of assembly method depends on the starting materials, and the products of each reaction determine which steps can follow.

Language models are commonly used to assist with individual steps of this process, including sequence analysis, primer design, and selection of cloning methods. Evaluating capabilities separately, however, leaves open whether models can successfully plan and execute a full experiment. For example, a proposed sequence may satisfy the requested design while an accompanying protocol may fail to produce it. Similarly, executable protocols may fail to address the scientist's objective. This planning process is time-consuming even for experienced scientists, with human errors frequent and surfacing only after several days (or weeks) of work. A model that can reliably reason through entire cloning workflows stands to provide enormous scientific and commercial benefit for scientists.

To measure progress toward this goal, we introduce BenchBench-Cloning, a benchmark of 65 molecular cloning tasks designed to evaluate a model's performance on correct and efficient cloning strategies. The protocol-development set contains 50 tasks where scientists specify the desired construct and the agent develops an executable protocol from the supplied inventory, and the construct-design set contains 15 tasks where the agent must design a construct that satisfies a scientific objective based on experimental motivations.

We contribute:
\begin{itemize}
    \item \textbf{BenchBench-Cloning:} a benchmark of 65 tasks testing AI capability in reasoning through complex cloning tasks.
    \item \textbf{Baseline results} for current frontier models, indicating that current models struggle to develop complete, valid cloning plans.
\end{itemize}

We describe the design and construction of the benchmark, present results from evaluating current models against it, and discuss what these results suggest about the current state and future trajectory of AI-assisted experimental planning.

\section{Related Work}

Life-science benchmarks increasingly evaluate tasks that resemble parts of a scientist's workflow. LAB-Bench and LABBench2 cover literature use, data interpretation, and sequence manipulation~\citep{laurent2024labbench,laurent2026labbench2}. LifeSciBench evaluates expert-authored research tasks using detailed rubrics across multiple scientific workflows~\citep{liu2026lifescibench}. These evaluations broaden the assessment of scientific assistance beyond factual recall to evidence interpretation and conclusion formulation. BenchBench-Protocol evaluates modifications to published laboratory protocols using tasks recovered from changes scientists made to experimental work~\citep{sivakumar2026benchbenchprotocol}.

Agentic benchmarks such as BixBench~\citep{mitchener2025bixbench} and Terminal-Bench-Science~\citep{tbscience2026} evaluate models on multistep computational analyses in biology and other scientific domains. LABBench2 CloningQA is the closest prior work, with 14 tasks requiring complete cloning protocols including methods like DNA and enzyme reagents, restriction-ligation cloning, Gibson assembly, and Golden Gate assembly. Submissions use a domain-specific language and are evaluated using component checks and end-to-end in silico validation~\citep{laurent2026labbench2}. BenchBench-Cloning expands this formulation to 50 protocol-development tasks in an environment supporting 42 operations. Our verification requirements include method-specific primer constraints, molecular-state and end-chemistry checks, and the ability to produce and isolate multiple products.

\section{Benchmark Design and Construction}
\label{sec:benchmark-design}

BenchBench-Cloning contains 65 tasks divided into two sets: 50 protocol-development tasks and 15 construct-design tasks. In protocol-development tasks, the desired construct is specified and the agent must determine how to produce it from the available materials. In construct-design, the scientific objective is specified and the agent must determine appropriate construct(s), specify their exact sequences, and provide a construction plan.

\begin{figure}[t]
\centering
\includegraphics[width=\linewidth]{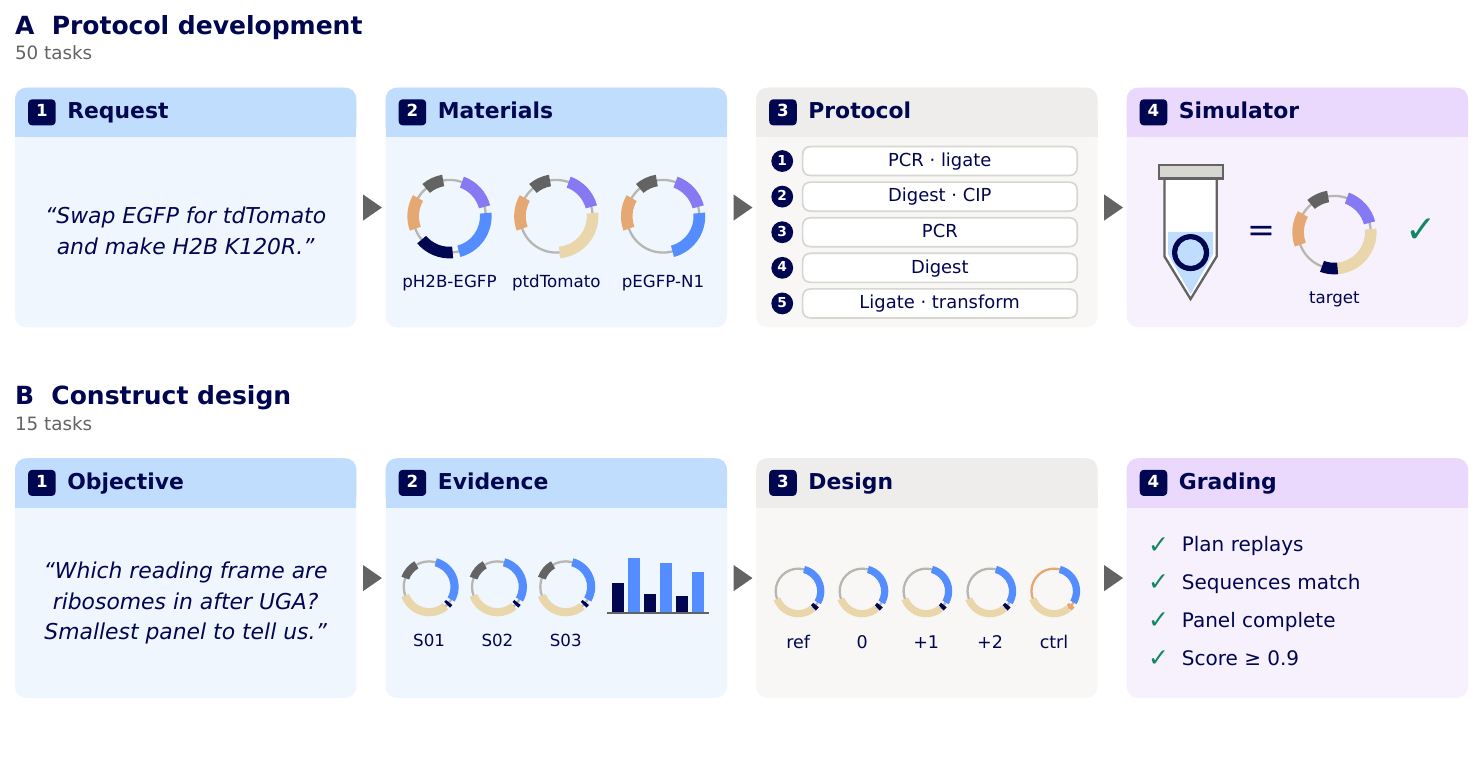}
\caption{The two task formats. In protocol development (A) the agent writes a protocol that the simulator executes; in construct design (B) the agent returns constructs and a plan that deterministic checks and, on ten tasks, a model judge score.}
\label{fig:task-overview}
\end{figure}

Both sets of tasks provided a fixed request and supporting files. Agents inspect these files, use analysis tools, and revise their work during an attempt. The two task sets differ in their inputs, outputs, and evaluation procedure (\figref{fig:task-overview}).

\subsection{Protocol-Development Tasks}
\label{sec:protocol-development}

Each protocol-development task provides a scientist's request describing the desired construct and an inventory of starting materials. Products are circular plasmids between 2.8kb and 15.2kb long, with a median length of 5.4kb. All 50 protocol-development tasks are derived from laboratory notebooks created by contributors on Benchling, drawing on real construct requests and experimental context. An inventory containing up to eight starting plasmid constructs is supplied as Genbank or FASTA files with accompanying sequence annotations. Notes of material properties (such as methylation sensitivity at restriction sites, GC-rich regions, end chemistry, acceptor vectors with ccdB counter-selection cassettes, host strain) are also provided. The task set spans nine experimental contexts, each defined by the dominant constraint the task imposes (\tabref{tab:task-families}).

\begin{table}[t]
\centering
\small
\begin{tabularx}{\linewidth}{>{\raggedright\arraybackslash}p{3.3cm}rrY}
\toprule
\textbf{Dominant constraint} & \textbf{Tasks} & \textbf{Share} & \textbf{Reasoning the task requires} \\
\midrule
Methylation state (unknown host / Dam-Dcm sites) & 11 & 22\% & Starting plasmid grown in an unknown or mixed host, or a required site overlapping a Dam/Dcm motif; establish a known methylation state or avoid the blocked site before cutting \\
\addlinespace
Homology-directed and seamless assembly & 7 & 14\% & Gibson, In-Fusion, in vivo assembly or recombineering with homology arms near the minimum length; in vitro transcription templates needing an oriented promoter \\
\addlinespace
Repeats and tandem arrays & 7 & 14\% & Repeated elements (tandem tags, repeated promoters or terminators, homopolymer tracts) where a primer or homology arm binds in more than one place or where a homopolymer run makes length unstable through slippage \\
\addlinespace
Backbone end chemistry and orientation & 6 & 12\% & Self-compatible vector ends that re-circularize or dimerize unless dephosphorylated; ends that must be filled or blunted; inserts whose blunt or palindromic ends give both orientations \\
\addlinespace
Golden Gate / MoClo & 5 & 10\% & Assemblies with internal Type IIS sites to domesticate and overhang sets that must stay unique \\
\addlinespace
GC-rich templates & 4 & 8\% & A window over the disclosed GC limit so PCR through it fails; the fragment must be released or built another way \\
\addlinespace
Recombination, CRISPR and viral vectors & 4 & 8\% & Gateway and Cre/lox recombination, CRISPR guide and donor plasmids, lentiviral and AAV transfer plasmids \\
\addlinespace
Counterselection & 4 & 8\% & An acceptor or destination vector carrying a cassette that the product must lose \\
\addlinespace
Expression vectors and tag exchange & 2 & 4\% & Bacterial expression vectors in which an affinity or solubility tag is exchanged while the reading frame and protease site are preserved \\
\midrule
\textbf{Total} & \textbf{50} & \textbf{100\%} & \\
\bottomrule
\end{tabularx}
\caption{Protocol-development task families. Task counts and shares are shown for each dominant constraint with the reasoning it requires.}
\label{tab:task-families}
\end{table}

The environment includes documentation for a custom cloning protocol language supporting 42 operations. These cover amplification, partial and complete digestion, Gibson and Golden Gate assembly, oligonucleotide annealing and ligation, end modification, CRISPR-based operations, and other relevant techniques. The environment also includes tools such as Python, Biopython, Primer3-py, and plotting libraries for inspection and design.

During evaluation, the agent receives the request, inventory, operation documentation, and analysis tools. It submits a protocol that is then passed through the simulator and assigned a binary score. A protocol passes if it successfully executes and ends with one isolated product that matches the hidden target sequence. Invalid operations, incorrect sequences, or unresolved mixtures result in failure. Alternative construction routes are accepted when they pass through the simulator and produce the same final product as the reference protocol.

\subsection{Construct-Design Tasks}
\label{sec:construct-design}

The construct-design set contains 15 tasks derived from published scientific studies.\footnote{Tasks, graders, and reference solutions for the construct-design set are available at \url{https://github.com/benchling/benchbench-cloning}.} Each task provides a scientist's objective alongside sequence records and supporting materials. Six tasks additionally provide experimental records such as sequencing reads or gel images. Agents return an annotated construct sequence, orderable primers when required, a construct plan, and a written justification.

These tasks require the agent to determine which constructs are needed to address a scientific objective. Depending on the task, a successful response may require identifying a relevant sequence feature, applying a repair across related constructs, or including controls needed to interpret an experiment.

Each task has a deterministic grader that checks the returned sequences, their source, and the construction route. Additionally, 10 tasks use a rubric-based LLM-as-a-judge to assess the written reasoning and construction plan. For these tasks, the final score is the product of the deterministic grader and the rubric score. Task sources are listed in \appref{app:construct-tasks}.

\subsection{Model Setup}
\label{sec:model-setup}

We evaluate eight systems using Claude Opus 5.5, Claude Mythos 5.1, GPT-6 Astra, GPT-6.1 Sol, Gemini 3.8 Flash, Gemini 4 Argon, Grok 4.7, and Kimi K3. Models were tested in their native agent harnesses where available, and Kimi K3 was tested using the Pi harness. Each system completed ten individual attempts per protocol-development task and five per construct-design task. \appref{app:model-config} shows the specific settings of each run.

Runs were orchestrated using Harbor in Modal sandboxes with two vCPUs and 6~GiB of memory. Network access was restricted to calls to the model provider. Each agent received the task instructions and supporting files, used the available tools, and wrote its submission to a designated answer file. Attempts were limited to two hours. GPT-6 Astra served as the model judge for the ten construct-design tasks requiring rubric-based evaluation. We also evaluated the tasks with Claude Mythos 5.1 and report agreement metrics in \appref{app:judge-agreement}.

\subsection{Evaluation Metrics}
\label{sec:metrics}

For the protocol-development tasks, we report pass rate, defined as the proportion of attempts whose submitted protocol executes successfully and produces the isolated target sequence. We average the ten attempt outcomes within each task and average across the fifty tasks. We estimate 95\% confidence intervals using 20,000 bootstrap resamples of the tasks, retaining all attempts associated with each sampled task. For construct-design tasks, an attempt is classified as successful when its final score is at least 0.9. The pass rate is computed by averaging the five attempt outcomes within each task and then averaging across the 15 tasks. The aggregate score is a weighted average of the protocol-development and construct-design scores, with weights of 50/65 and 15/65, respectively, giving each task equal weight.

For protocol-development tasks, we additionally report \passatk{} for up to ten attempts to measure the probability that at least one of $k$ attempts succeeds. This measures the availability of a valid protocol among repeated attempts.

\section{Results}
\label{sec:results}

\begin{figure}[t]
\centering
\includegraphics[width=\linewidth]{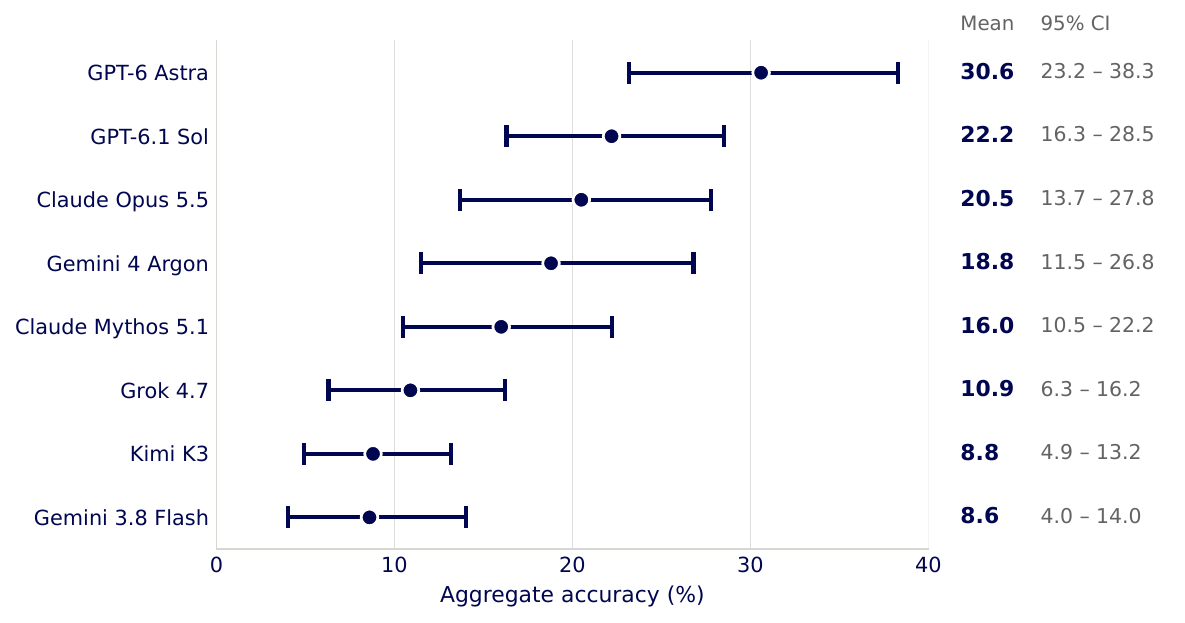}
\caption{Aggregate score by system. Task-weighted means at top effort are shown with 95\% confidence intervals.}
\label{fig:overall-performance}
\end{figure}

Performance on BenchBench-Cloning varies substantially across systems. GPT-6 Astra is the strongest overall system, with an aggregate score of 30.6\%. Scores for the other evaluated systems range from 8.6\% to 22.2\% (\figref{fig:overall-performance}).

\subsection{Performance by Task Set}

\begin{figure}[t]
\centering
\includegraphics[width=\linewidth]{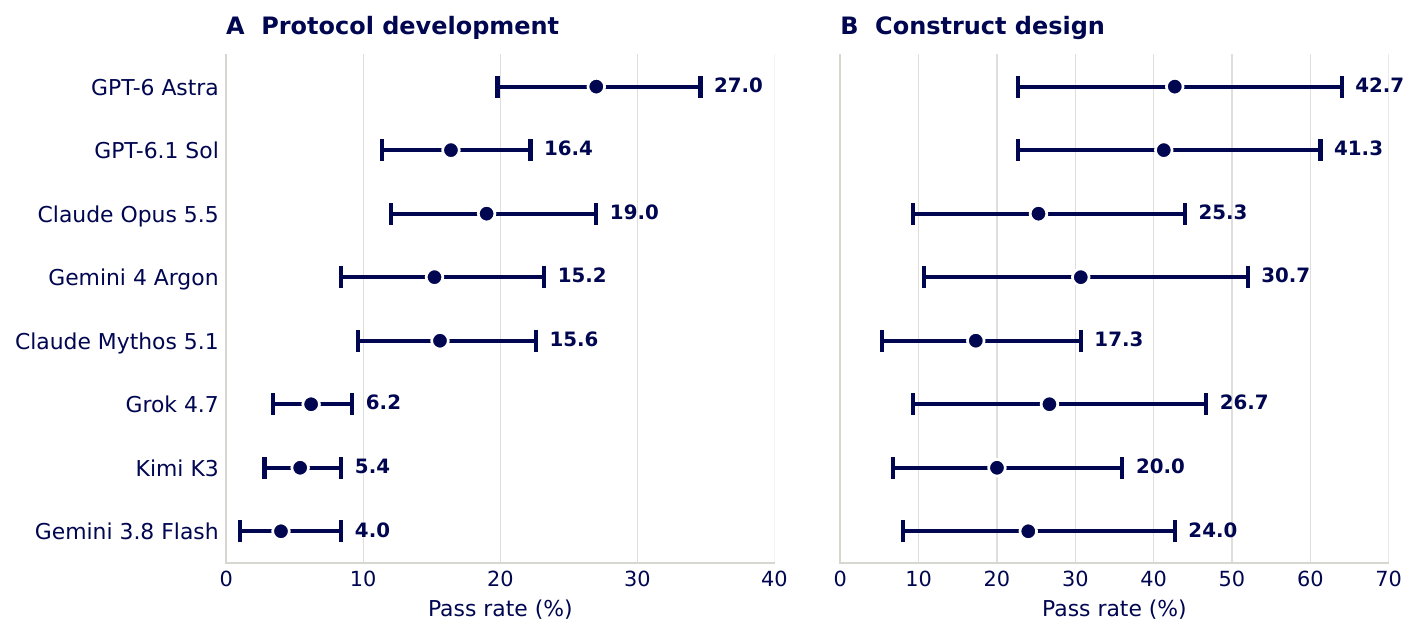}
\caption{Pass rate by task set. Means at top effort are shown with 95\% confidence intervals for protocol development (A) and construct design (B).}
\label{fig:task-set-performance}
\end{figure}

On the protocol-development set, GPT-6 Astra achieves the highest pass rate at 27.0\% (95\% CI 19.8--34.6\%), followed by Claude Opus 5.5 at 19.0\% and GPT-6.1 Sol at 16.4\% (\figref{fig:task-set-performance}). Grok 4.7, Kimi K3, and Gemini 3.8 Flash score 6.2\%, 5.4\%, and 4.0\%, respectively (\tabref{tab:results}).

\begin{table}[t]
\centering
\small
\begin{adjustbox}{max width=\linewidth}
\begin{tabular}{llccc}
\toprule
\textbf{System} & \textbf{Effort} & \textbf{Protocol dev.\ \%} & \textbf{Construct design \%} & \textbf{Aggregate \%} \\
\midrule
GPT-6 Astra & max & 27.0 [19.8, 34.6] & 42.7 [22.7, 64.0] & 30.6 [23.2, 38.3] \\
GPT-6.1 Sol & max & 16.4 [11.4, 22.2] & 41.3 [22.7, 61.3] & 22.2 [16.3, 28.5] \\
Claude Opus 5.5 & max & 19.0 [12.0, 27.0] & 25.3 [9.3, 44.0] & 20.5 [13.7, 27.8] \\
Gemini 4 Argon & high & 15.2 [8.4, 23.2] & 30.7 [10.7, 52.0] & 18.8 [11.5, 26.8] \\
Claude Mythos 5.1 & max & 15.6 [9.6, 22.6] & 17.3 [5.3, 30.7] & 16.0 [10.5, 22.2] \\
Grok 4.7 & xhigh & 6.2 [3.4, 9.2] & 26.7 [9.3, 46.7] & 10.9 [6.3, 16.2] \\
Kimi K3 & high & 5.4 [2.8, 8.4] & 20.0 [6.7, 36.0] & 8.8 [4.9, 13.2] \\
Gemini 3.8 Flash & high & 4.0 [1.0, 8.4] & 24.0 [8.0, 42.7] & 8.6 [4.0, 14.0] \\
\bottomrule
\end{tabular}
\end{adjustbox}
\caption{Performance by task set at top effort. Pass rates in percent are shown with 95\% confidence intervals.}
\label{tab:results}
\end{table}

All evaluated systems produce successful protocols, but no system consistently completes the protocol-development tasks. Analysis of submitted designs shows that unsuccessful attempts can contain substantial correct work, with a small number of decisions preventing completion.

\subsection{Performance Under Repeated Sampling}

\begin{figure}[t]
\centering
\includegraphics[width=0.9\linewidth]{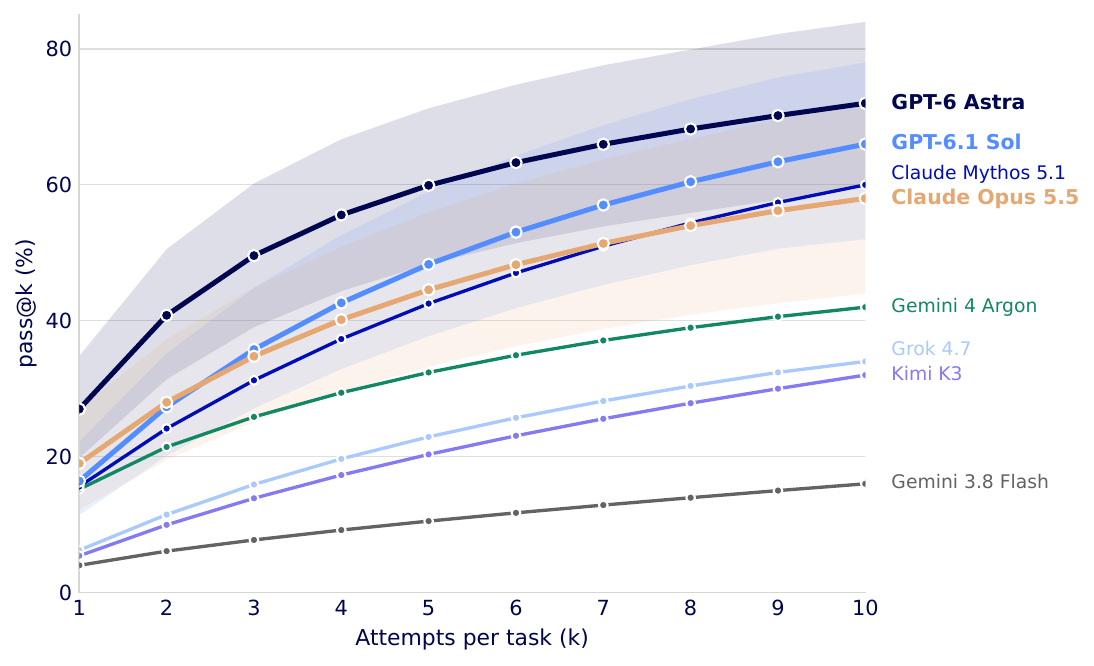}
\caption{Protocol-development \passatk{}. Probability that at least one of $k$ attempts passes is shown with 95\% confidence bands for the three leading systems.}
\label{fig:pass-at-k}
\end{figure}

Because a scientist can compare several candidate answers before acting on one, expected \bestatk{} performance is reported as an upper bound on single-system performance. System performance ranking is not stable in $k$ (\figref{fig:pass-at-k}). Opus performs better than Sol on individual attempts, but Sol performs better at $k=10$. Opus's successful attempts are concentrated on fewer tasks, while Sol succeeds less consistently across a broader set of tasks.

At ten attempts, Astra solves 36 of the 50 protocol-development tasks at least once, corresponding to a pass@10 of 72\% (\tabref{tab:pass-at-k}). Sol solves 33 tasks and Opus solves 29, corresponding to 66\% and 58\%, respectively. The strongest system still leaves 14 tasks unsolved, indicating that the benchmark is not saturated within ten attempts.

\subsection{Reasoning Effort and Cost}

\begin{figure}[t]
\centering
\includegraphics[width=0.95\linewidth]{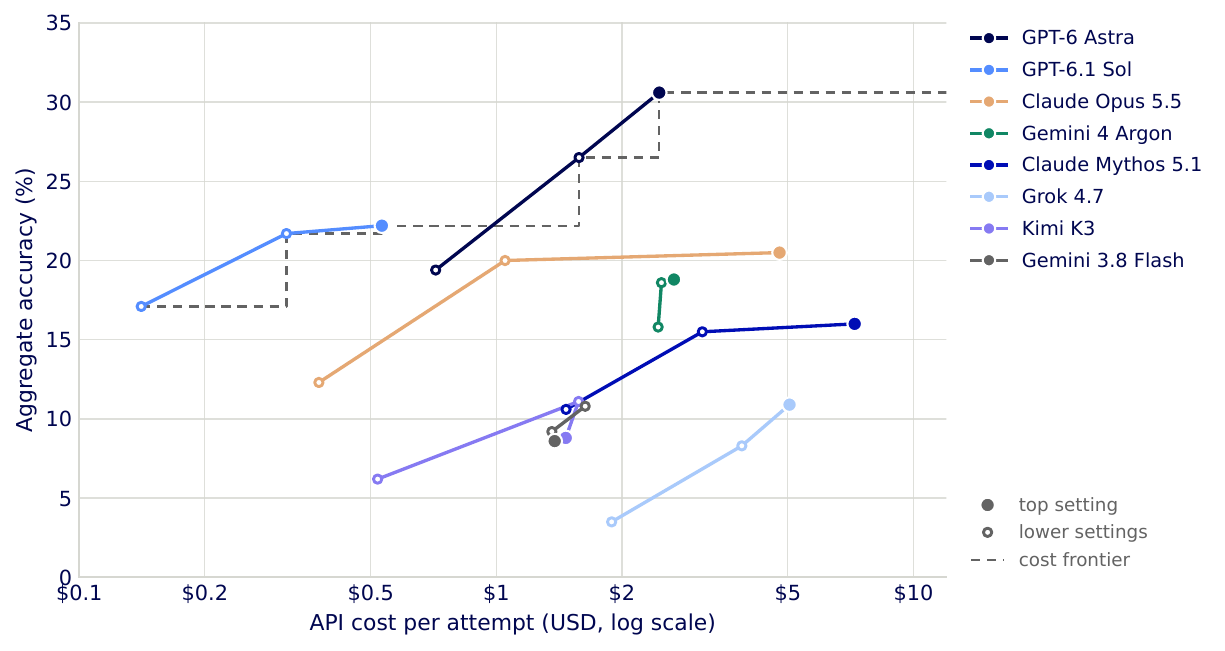}
\caption{Aggregate score against cost per attempt. Each system is shown at its three reasoning settings with the top setting filled; the dashed line is the cost frontier.}
\label{fig:accuracy-vs-cost}
\end{figure}

Average cost per attempt varies by more than an order of magnitude across systems at comparable performance (\figref{fig:accuracy-vs-cost}). The pareto frontier identifies the system that offers the best observed trade-off between performance and cost. Token counts for each run are reported in \appref{app:cost}.

\section{Analysis of Model Behavior}
\label{sec:analysis}

\subsection{Successful Work}

\begin{figure}[t]
\centering
\includegraphics[width=\linewidth]{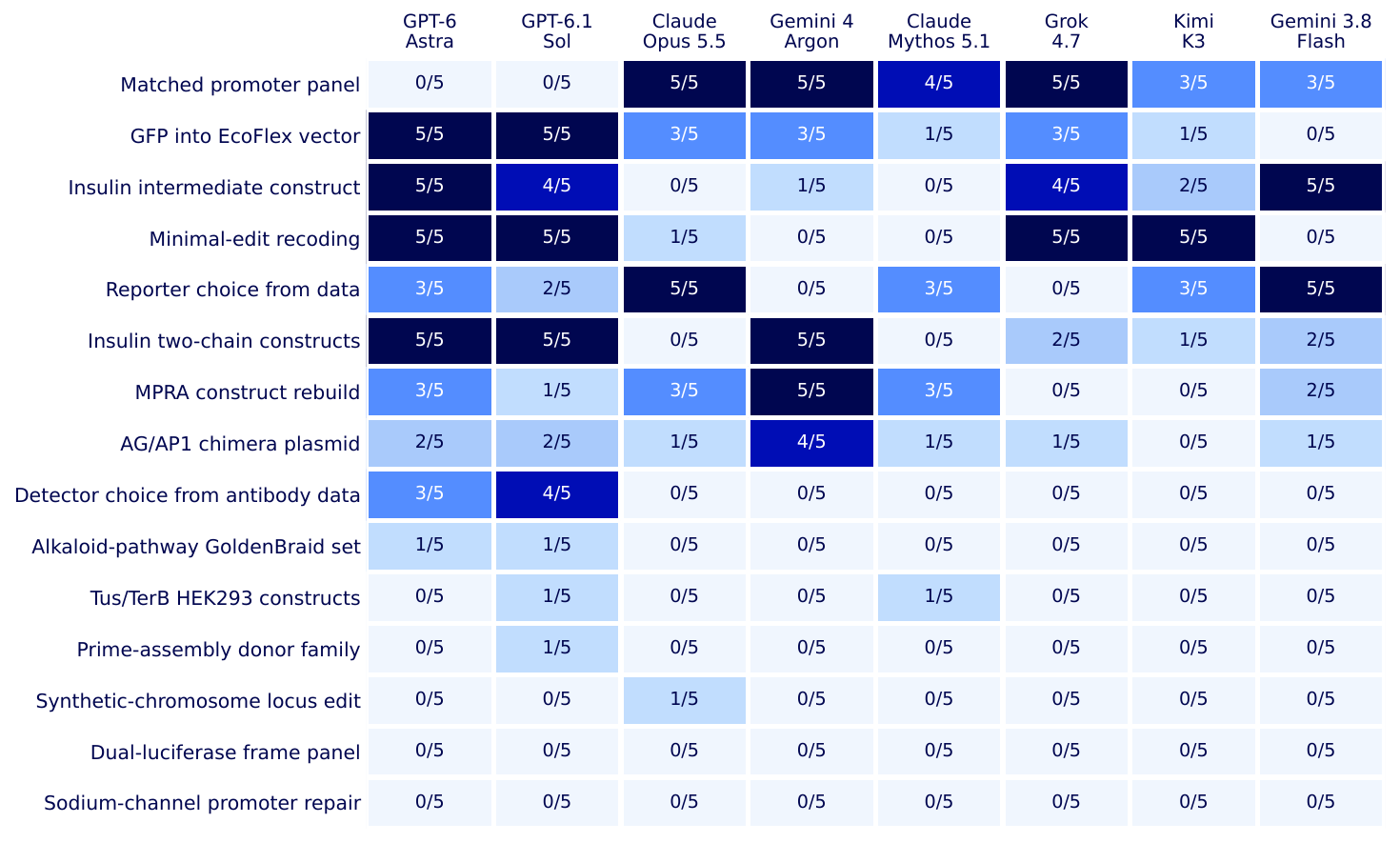}
\caption{Construct-design solves by task and system. Cells give the number of five attempts that passed at top effort.}
\label{fig:construct-by-task}
\end{figure}

Systems perform well on several tasks where the expected molecular outcome is spelled out clearly. Six of the fifteen construct-design tasks were passed in every attempt by at least two systems, and four systems passed the minimal-edit recoding task every time (\figref{fig:construct-by-task}; \tabref{tab:construct-grid}). Many failed submissions also contained correct work. In the sodium-channel task, every scored attempt placed both introns at the junctions and kept the full channel protein intact, even though none of them passed the task.

\begin{figure}[t]
\centering
\includegraphics[width=\linewidth]{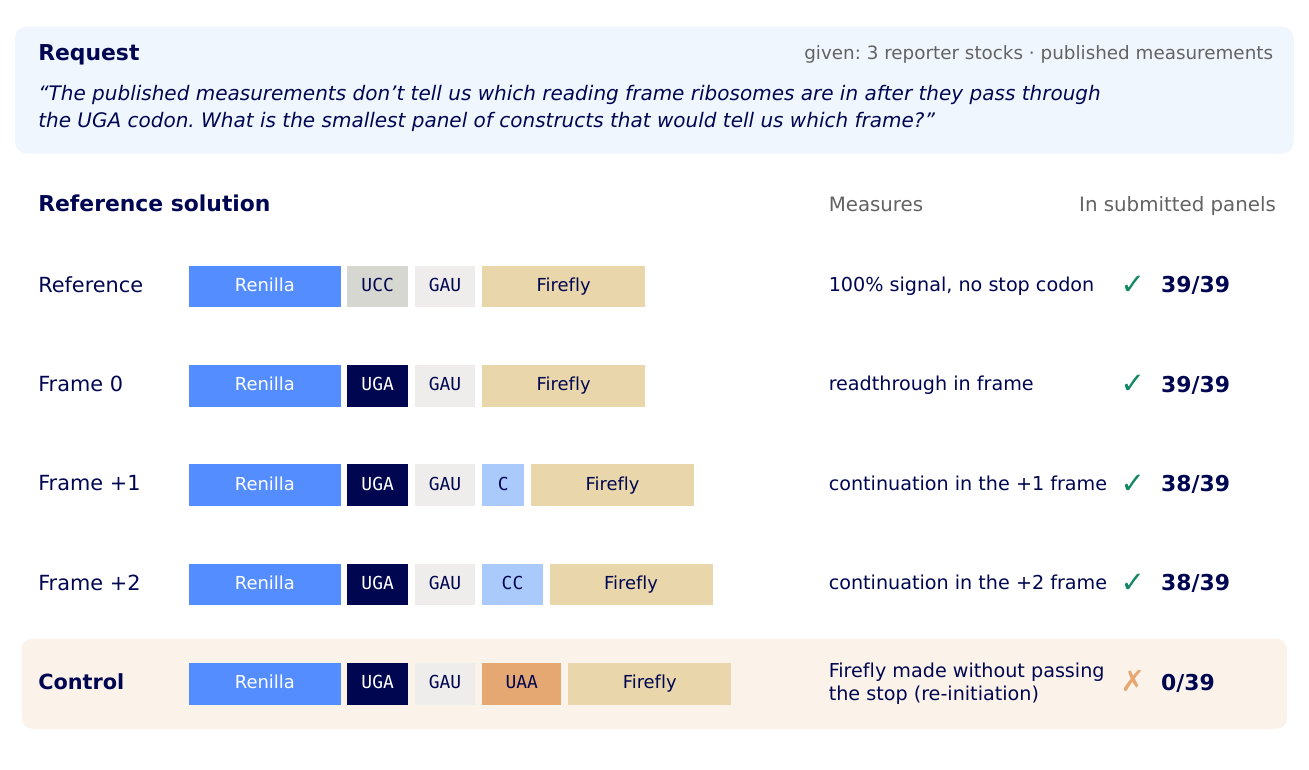}
\caption{The dual-luciferase frame-panel task. The request, the five reference constructs, and the number of the 39 submitted panels containing each construct are shown.}
\label{fig:dual-luciferase}
\end{figure}

The dual-luciferase task makes this especially clear. Across the eight systems, 38 of the 39 submitted panels had correctly built reference, frame 0, frame $+1$ and frame $+2$ constructs, and every construction step in those panels ran as written. None of them included the fifth construct, a control with the Firefly reading frame closed by a stop codon, which the experiment needs in order to separate readthrough from re-initiation (\figref{fig:dual-luciferase}). Several systems even mentioned re-initiation in their written answers, but they treated it as background shared by all constructs rather than something the panel had to measure.

Taken together, these results suggest that agents can usually carry out the individual sequence edits a task requires. Where they fall short is in reasoning from the experimental question to the set of constructs required to exclude alternative explanations.

\subsection{Protocol-Development Failure Modes}

Analysis of protocol failures identifies recurring errors in primer design, backbone preparation, and methylation state. These errors often arise in systems that correctly identify the intended sequence changes but do not maintain compatibility between the materials and operations used to produce them.

In several cases, we found that systems would frequently select primers which target the correct sequence while remaining unsuitable for the proposed reaction conditions. Primer failures frequently involve pairs that cannot be used together under the simulated annealing conditions. For example, primer pairs were found to dimerize, or to have annealing points which were too far apart to be feasibly used in a reaction together. In these cases, while the system succeeded in designing single appropriate primers, it failed to evaluate whether the primer could be appropriately used as a pair. We also identified primer design failures involving secondary structure issues (e.g.\ hairpin loops), length limits, or homopolymer runs.

Backbone preparation and methylation produced additional failures. Systems ligated vectors with self-compatible ends without dephosphorylating them, leading to backbones recircularizing or dimerizing. In other cases, systems treated compatible or blunt ends as directional, carrying inserts forward in one orientation when the reaction yields both, or ignored methylation-sensitive sites in plasmids from a Dam/Dcm-methylating host, leaving the acceptor circular. These are mistakes a molecular biologist would ordinarily anticipate and prevent when designing the protocol. Each attempt is classified by the first error that blocks execution, reported by system in \appref{app:failures}.

\subsection{Construct-Design Failure Modes}

Most construct-design failures were incomplete designs. Systems usually made the change that was asked for, but they often missed a second change that the experiment needed. The sodium-channel task is a good example (\tabref{tab:construct-grid}). Two related expression constructs each carry a cryptic bacterial promoter that has to be removed. GPT-6 Astra and GPT-6.1 Sol fixed the first construct in all of their attempts, yet no system ever fixed the second one. The dual-luciferase task shows a similar pattern. Nearly all submitted panels had correct reference and frame-shifted reporters, but none of them included the closed-frame control.

Some designs went further than the task asked. The matched-panel task asks for the smallest set of constructs that can resolve a single discrepancy, and the grader checks for this. Astra and Sol built an eight-construct factorial panel in every attempt and so failed the check. Most of the other systems submitted the four-construct panel and passed.

When the grader also checked the construction route, many failures came from the construction record and not from the sequences themselves. We saw source references that could not be resolved, the same identifier used for different sequences, missing oligonucleotide tables, and fragments that no listed reaction actually produces. The designed sequences in these submissions were sometimes correct, but the grader had no way to confirm that the plan could build them from the materials provided.

Ten of the tasks also use a model judge, so passing the deterministic checks does not guarantee a pass. The judge scores the written plan, and an attempt needs at least 0.9 to pass. At the top effort setting, 34 of the 142 attempts that cleared the deterministic checks fell below this threshold (\tabref{tab:judge-gate}). Most often the judge's main concern was that the plan did not lay out an orderable path from the supplied sources to the final product; in two cases it was a misreading of the experimental evidence. We list the reviewed submissions in \appref{app:reviewed}.

\section{Discussion}
\label{sec:discussion}

\subsection{Implications and Future Work}

BenchBench-Cloning shows that frontier models can simulate many of the individual operations required by molecular cloning, but they do not yet reliably assemble those operations into a complete, executable experimental design. The strongest system, GPT-6 Astra, reaches an aggregate score of 30.6\% and a protocol-development pass rate of 27.0\%, and even at ten attempts it leaves 14 of 50 protocol-development tasks unsolved. However, it comes with significant costs; under the pareto frontier, it is observed that mid-range systems like GPT-6.1 Sol can achieve similar results at lower cost. Notably, while repeated sampling produces a greater likelihood of valid solutions, its practical value depends on identifying those solutions. Comparing individual attempts with multiple candidates and a selection procedure under the same inference budget would help determine how additional computation is best used.

Failures are rarely a matter of missing basic knowledge. Unsuccessful submissions are often the result of flawed critical thinking and understanding of the deeper underlying motivations, such as an omitted control. The failures are not at any single subproblem, but rather in the ability to maintain context across longer horizon tasks, such as understanding that materials have constraints, or that a primer pair cannot work together under the stated reaction conditions.

These findings bear on how such systems are used today. A failing submission typically reads as a complete and well-reasoned plan, with correct sequence edits, plausible primers, and a confident justification, and any mistakes are only visible to someone who already knows what the experiment demands. Current systems are therefore convincing enough to mislead the users most likely to rely on them. The near-miss character of these failures also points to the fix, with blocking errors being checks an experienced molecular biologist makes as a matter of course that a system should make on its own plan before presenting it. Until systems do so reliably, expert review of model-generated protocols remains necessary; the benchmark measures progress toward systems that reliably and correctly plan complete cloning workflows. Further analysis can assess whether explicit self-verification and candidate selection improve complete-task success.

For benchmark design, our results also show that grading choices matter: of 142 submissions that passed the deterministic gate, 34 fell below the pass threshold on the model judge, most often because the plan lacked an orderable construction route from the supplied sources. Correct sequences thus do not guarantee a usable plan, and some failures reflect incomplete documentation rather than flawed biology.

\subsection{Limitations}

While BenchBench-Cloning was designed as a single-turn benchmark evaluating whether a model could respond correctly to a fixed scientist request, this likely does not reflect how a scientist would collaborate with a model. Scientists often use LLMs in a multi-turn capacity, which would allow models to ask clarifying questions about available materials, revise plans after feedback, respond to experimental results obtained during construction, and emphasize practical limitations on suggested protocols. Future work can focus on a format that more closely reflects this reality.

Further, while BenchBench-Cloning measures planning for molecular cloning, it does not directly measure whether models improve experimental outcomes at the bench. The simulator cannot capture every source of variation in reagents, handling, growth, or reaction outcomes. As LLMs move into the physical world, future iterations of BenchBench-Cloning could involve actually conducting experiments designed by the models to test out the protocols designed.

Another limitation in analysis of traces is that our simulator inherently reports the first blocking error, so these diagnostics do not necessarily establish whether later steps would have succeeded. Failure frequencies also depend on the routes models choose. Models using PCR more frequently encounter primer-related failures, which may have skewed the occurrences of primer design failures amongst the reasoning traces.

Finally, a limitation of our construct-design tasks was the use of LLM-as-a-judge to assess written reasoning and construction plans. Because this kind of model grading is inherently stochastic, there were occasional inconsistencies in grading results. Notably, agreement between graders does not establish independent scientific validity.

\section{Conclusion}

BenchBench-Cloning measures whether models can determine the constructs addressing a scientific objective and plan a construction route from the available materials. GPT-6 Astra scores highest at 30.6\%, with other systems' aggregate scores ranging from 8.6\% to 22.2\%. Repeated sampling improves protocol-development performance, but the strongest system still leaves 14 out of 50 tasks unsolved after ten attempts.

A closer look at successes and failures identified trends. Systems demonstrate surprisingly skilled sequence work and successful construction planning on part of the benchmark. However, their failures frequently concern compatibility between operations, incomplete construct plans, or lack of consideration towards any changes needed for a broader scientific objective. These failures matter in practice because the scientists most likely to rely on a system for cloning judgment are those with the least bench experience, and the least able to catch its errors. Evaluating molecular cloning assistance therefore requires checking both whether a design addresses the experiment and whether the proposed operations can produce it. Closing the gap between individual cloning decisions and a complete experimental plan is the next step toward systems that scientists can rely on, and BenchBench-Cloning offers a way to measure progress toward it.

% \subsubsection*{Acknowledgments}
% Add acknowledgments here before posting.

\clearpage
\bibliographystyle{iclr2026_conference}
\bibliography{references}

\clearpage
\appendix
\makeatletter
\@addtoreset{figure}{section}
\@addtoreset{table}{section}
\makeatother
\renewcommand{\thefigure}{\Alph{section}\arabic{figure}}
\renewcommand{\thetable}{\Alph{section}\arabic{table}}

\section{Construct-Design Tasks and Sources}
\label{app:construct-tasks}

Each construct-design task gives the agent a scientist's objective, sequence records and supporting materials. Six tasks also supply experimental records from the source study. Every task has a deterministic grader that checks the returned sequences, their sources and the construction route; ten tasks also use a rubric-based model judge to assess the written reasoning and construction plan, and for these the final score is the product of the deterministic result and the rubric score. An attempt passes at a final score of at least 0.9. \tabref{tab:construct-tasks} lists the tasks; \figrefs{fig:task-overview}{fig:dual-luciferase} show the dual-luciferase task in detail.

\begin{table}[h]
\centering
\scriptsize
\begin{tabularx}{\linewidth}{rYYYl}
\toprule
\textbf{\#} & \textbf{Task} & \textbf{Source} & \textbf{Experimental records supplied} & \textbf{Grading} \\
\midrule
1 & MPRA construct rebuild & Dao et al.\ 2025, \textit{Nat.\ Commun.} & Sequencing reads from the plasmid pool and cellular RNA & Deterministic \\
2 & Matched promoter panel & Pic\'o et al.\ 2026, \textit{Nat.\ Commun.} & Per-well growth and GFP measurements, part map & Deterministic \\
3 & Sodium-channel promoter repair & \textit{J.\ Biol.\ Chem.} 2021, cryptic prokaryotic promoters in neuronal sodium channels & Gel and fragment-stability images, sequence changes, start-site and electrophysiology data & Deterministic \\
4 & Detector choice from antibody data & \textit{PLoS ONE} 2021, CD55 C81A reporter & Antibody and localisation images, oligonucleotide and genotype tables & Deterministic + rubric judge \\
5 & Reporter choice from data & Ghodke et al.\ 2016, \textit{Biophys.\ J.} & Published microscopy and calibration figure panels & Deterministic \\
6 & Dual-luciferase frame panel & Annibaldis et al.\ 2020, \textit{Nucleic Acids Res.} & Published readthrough measurements (three panels) & Deterministic \\
7 & AG/AP1 chimera plasmid & Lai et al.\ 2021, \textit{Nat.\ Commun.} & Source paper and supplement & Deterministic + rubric judge \\
8 & Synthetic-chromosome locus edit & Goold et al.\ 2025, \textit{Nat.\ Commun.} & None (genome and design records) & Deterministic + rubric judge \\
9 & Alkaloid-pathway GoldenBraid set & Grzech et al.\ 2023, \textit{ACS Synth.\ Biol.} & Source paper and supplement & Deterministic + rubric judge \\
10 & GFP into EcoFlex vector & EcoFlex kit (Addgene) & None & Deterministic + rubric judge \\
11 & Prime-assembly donor family & Liu et al.\ 2025, \textit{bioRxiv} & Source preprint and supplement & Deterministic + rubric judge \\
12 & Tus/TerB HEK293 constructs & Addgene plasmids and genomic records & None & Deterministic + rubric judge \\
13 & Minimal-edit recoding & Authored sequence & None & Deterministic + rubric judge \\
14 & Insulin two-chain constructs & Human INS and pMAL-c6T records & None & Deterministic + rubric judge \\
15 & Insulin intermediate construct & Human INS and pMAL-c6T records & None & Deterministic + rubric judge \\
\bottomrule
\end{tabularx}
\caption{Construct-design tasks. Source, experimental records supplied, and grading route are listed for each task.}
\label{tab:construct-tasks}
\end{table}

\clearpage
\section{Model and Harness Settings}
\label{app:model-config}

All attempts ran through Harbor in Modal sandboxes (2 vCPU, 6~GiB) with network access limited to the model provider. Each system completed 10 attempts per protocol-development task and 5 attempts per construct-design task at every reasoning setting.

\begin{table}[h]
\centering
\scriptsize
\begin{tabularx}{\linewidth}{lllYl}
\toprule
\textbf{System} & \textbf{Model id} & \textbf{Harness (version)} & \textbf{Access route} & \textbf{Reasoning settings} \\
\midrule
GPT-6 Astra & \texttt{gpt-6-astra} & Codex CLI 0.156.1 & OpenAI API & low / high / max \\
GPT-6.1 Sol & \texttt{gpt-6.1-sol} & Codex CLI 0.156.1 & OpenAI API & low / high / max \\
Claude Opus 5.5 & \texttt{claude-opus-5-5} & Claude Code 2.1.289 & Anthropic API (WIF) & low / high / max \\
Gemini 4 Argon & \texttt{gemini-4-argon} & Gemini CLI 0.63.0, in-sandbox proxy & Vertex AI (global endpoint) & low / medium / high \\
Claude Mythos 5.1 & \texttt{claude-mythos-5-1} & Claude Code 2.1.289 & Anthropic API (WIF) & low / high / max \\
Grok 4.7 & \texttt{grok-4.7} & Grok Build 1.0.46 & xAI API & low / high / xhigh \\
Kimi K3 & \texttt{moonshotai/Kimi-K3} & Pi 1.0.3 & Baseten & low / medium / high \\
Gemini 3.8 Flash & \texttt{gemini-3.8-flash} & Gemini CLI 0.63.0 & Vertex AI (API key) & low / medium / high \\
\bottomrule
\end{tabularx}
\caption{Model and harness configurations. Model identifier, harness, access route, and reasoning settings are listed per system.}
\label{tab:model-config}
\end{table}

\section{Judge Agreement}
\label{app:judge-agreement}

To check the rubric-based outcomes, every construct-design attempt from the first six systems evaluated was graded a second time by Claude Mythos 5.1 acting as an independent judge. The judge received the public task text, the private reference material and the submission files, and returned a score on the same 0 to 1 scale; agreement is measured on the binary pass outcome.

\begin{table}[h]
\centering
\small
\begin{tabular}{lrrr}
\toprule
\textbf{Slice} & \textbf{Trials graded} & \textbf{Agreement} & \textbf{Cohen's kappa} \\
\midrule
All trials & 387 & 89.7\% & 0.76 \\
Tasks with experimental records (6) & 137 & 99.3\% & 0.98 \\
Construction tasks (9) & 250 & 84.4\% & 0.65 \\
\midrule
GPT-6 Astra & 62 & 85.5\% & 0.72 \\
GPT-6.1 Sol & 63 & 84.1\% & 0.69 \\
Claude Opus 5.5 & 74 & 85.1\% & 0.67 \\
Kimi K3 & 62 & 96.8\% & 0.90 \\
Gemini 3.8 Flash & 63 & 95.2\% & 0.80 \\
Grok 4.7 & 63 & 92.1\% & 0.79 \\
\bottomrule
\end{tabular}
\caption{Agreement between the Claude Mythos 5.1 judge and reported outcomes. Trials graded, agreement, and Cohen's kappa are shown per slice.}
\label{tab:judge-agreement}
\end{table}

Agreement between Claude Mythos 5.1, used as an independent judge, and the outcomes reported in the paper (pass = final score of at least 0.9 under both). The comparison covers the 450 construct-design attempts of the first six systems evaluated (15 tasks, 5 attempts each). The Mythos judge saw the public task text, the private reference material and the submission files, but not the reported outcome. It declined to grade 63 attempts under its own safety policy; those are excluded.

\clearpage
\section{Token Usage and Cost}
\label{app:cost}

\begin{table}[h]
\centering
\scriptsize
\begin{adjustbox}{max width=\linewidth}
\begin{tabular}{llrrrrrrr}
\toprule
\textbf{System} & \textbf{Effort} & \textbf{Input (k)} & \textbf{Cached (k)} & \textbf{Output (k)} & \textbf{Cost/attempt (\$)} & \textbf{Median min} & \textbf{Cap hits} & \textbf{Design cost (\$)} \\
\midrule
GPT-6 Astra & low & 142 & 118 & 3 & 0.57 & 1.1 & 0 & 1.20 \\
GPT-6 Astra & high & 346 & 307 & 11 & 1.35 & 3.8 & 0 & 2.34 \\
GPT-6 Astra & max & 428 & 376 & 21 & 2.08 & 7.9 & 0 & 3.72 \\
GPT-6.1 Sol & low & 195 & 171 & 4 & 0.12 & 1.4 & 0 & 0.21 \\
GPT-6.1 Sol & high & 392 & 354 & 14 & 0.27 & 4.5 & 0 & 0.46 \\
GPT-6.1 Sol & max & 499 & 444 & 28 & 0.46 & 14.0 & 13 & 0.77 \\
Claude Opus 5.5 & low & 243 & 225 & 8 & 0.29 & 1.4 & 0 & 0.66 \\
Claude Opus 5.5 & high & 541 & 499 & 23 & 0.77 & 4.0 & 0 & 1.98 \\
Claude Opus 5.5 & max & 3,561 & 3,321 & 139 & 4.69 & 24.2 & 18 & 5.06 \\
Gemini 4 Argon & low & 5,076 & 4,620 & 39 & 2.21 & 6.5 & 1 & 3.22 \\
Gemini 4 Argon & medium & 4,184 & 3,760 & 74 & 2.53 & 10.6 & 2 & 2.35 \\
Gemini 4 Argon & high & 4,150 & 3,640 & 80 & 2.73 & 11.1 & 1 & 2.45 \\
Claude Mythos 5.1 & low & 299 & 269 & 14 & 1.13 & 2.9 & 0 & 2.60 \\
Claude Mythos 5.1 & high & 597 & 529 & 37 & 2.82 & 7.1 & 0 & 4.11 \\
Claude Mythos 5.1 & max & 1,224 & 1,050 & 96 & 7.23 & 18.8 & 3 & 7.23 \\
Grok 4.7 & low & 2,651 & 2,432 & 54 & 1.98 & 14.8 & 0 & 1.59 \\
Grok 4.7 & high & 4,658 & 4,251 & 165 & 4.00 & 42.3 & 15 & 3.47 \\
Grok 4.7 & xhigh & 6,184 & 5,598 & 199 & 5.30 & 43.3 & 30 & 4.20 \\
Kimi K3 & low & 239 & 185 & 21 & 0.54 & 4.0 & 2 & 0.45 \\
Kimi K3 & medium & 691 & 587 & 62 & 1.42 & 10.2 & 4 & 2.09 \\
Kimi K3 & high & 682 & 561 & 64 & 1.49 & 12.8 & 2 & 1.39 \\
Gemini 3.8 Flash & low & 10,959 & 10,240 & 80 & 1.61 & 17.7 & 37 & 1.71 \\
Gemini 3.8 Flash & medium & 7,148 & 6,610 & 109 & 1.31 & 17.5 & 13 & 1.52 \\
Gemini 3.8 Flash & high & 6,862 & 6,313 & 107 & 1.29 & 17.8 & 23 & 1.68 \\
\bottomrule
\end{tabular}
\end{adjustbox}
\caption{Token usage, cost, and runtime per protocol-development attempt. Mean tokens, cost, median minutes, and time-cap hits are shown per lane with the construct-design cost per attempt.}
\label{tab:token-usage}
\end{table}

\begin{table}[h]
\centering
\scriptsize
\begin{adjustbox}{max width=\linewidth}
\begin{tabular}{lllcccrr}
\toprule
\textbf{System} & \textbf{Tier} & \textbf{Effort} & \textbf{Protocol dev.\ \% [95\% CI]} & \textbf{Construct design \% [95\% CI]} & \textbf{Aggregate \% [95\% CI]} & \textbf{Proto (\$)} & \textbf{Const (\$)} \\
\midrule
GPT-6 Astra & low & low & 16.8 [10.2, 24.0] & 28.0 [13.3, 45.3] & 19.4 [13.2, 26.2] & 0.57 & 1.20 \\
GPT-6 Astra & mid & high & 21.2 [14.8, 28.6] & 44.0 [22.7, 66.7] & 26.5 [19.4, 34.0] & 1.35 & 2.34 \\
GPT-6 Astra & top & max & 27.0 [19.8, 34.6] & 42.7 [22.7, 64.0] & 30.6 [23.2, 38.3] & 2.08 & 3.72 \\
GPT-6.1 Sol & low & low & 15.4 [9.0, 22.6] & 22.7 [9.3, 38.7] & 17.1 [11.2, 23.5] & 0.12 & 0.21 \\
GPT-6.1 Sol & mid & high & 17.0 [11.4, 23.4] & 37.3 [20.0, 56.0] & 21.7 [15.7, 28.0] & 0.27 & 0.46 \\
GPT-6.1 Sol & top & max & 16.4 [11.4, 22.2] & 41.3 [22.7, 61.3] & 22.2 [16.3, 28.5] & 0.46 & 0.77 \\
Claude Opus 5.5 & low & low & 7.2 [3.8, 11.2] & 29.3 [13.3, 46.7] & 12.3 [7.7, 17.2] & 0.29 & 0.66 \\
Claude Opus 5.5 & mid & high & 15.6 [10.4, 21.6] & 34.7 [17.3, 53.3] & 20.0 [14.2, 26.3] & 0.77 & 1.98 \\
Claude Opus 5.5 & top & max & 19.0 [12.0, 27.0] & 25.3 [9.3, 44.0] & 20.5 [13.7, 27.8] & 4.69 & 5.06 \\
Gemini 4 Argon & low & low & 11.0 [7.4, 15.0] & 31.9 [10.7, 52.0] & 15.8 [10.4, 21.7] & 2.21 & 3.22 \\
Gemini 4 Argon & mid & medium & 14.2 [9.0, 20.0] & 33.3 [13.3, 54.7] & 18.6 [12.3, 25.4] & 2.53 & 2.35 \\
Gemini 4 Argon & top & high & 15.2 [8.4, 23.2] & 30.7 [10.7, 52.0] & 18.8 [11.5, 26.8] & 2.73 & 2.45 \\
Claude Mythos 5.1 & low & low & 9.0 [5.0, 13.8] & 16.0 [4.0, 29.3] & 10.6 [6.3, 15.4] & 1.13 & 2.60 \\
Claude Mythos 5.1 & mid & high & 13.0 [7.6, 19.2] & 24.0 [8.0, 42.7] & 15.5 [9.8, 21.8] & 2.82 & 4.11 \\
Claude Mythos 5.1 & top & max & 15.6 [9.6, 22.6] & 17.3 [5.3, 30.7] & 16.0 [10.5, 22.2] & 7.23 & 7.23 \\
Grok 4.7 & low & low & 1.4 [0.4, 2.8] & 10.7 [2.7, 21.3] & 3.5 [1.4, 6.2] & 1.98 & 1.59 \\
Grok 4.7 & mid & high & 4.0 [2.0, 6.4] & 22.7 [6.7, 41.3] & 8.3 [4.3, 12.9] & 4.00 & 3.47 \\
Grok 4.7 & top & xhigh & 6.2 [3.4, 9.2] & 26.7 [9.3, 46.7] & 10.9 [6.3, 16.2] & 5.30 & 4.20 \\
Kimi K3 & low & low & 3.2 [1.2, 6.0] & 16.0 [2.7, 33.3] & 6.2 [2.5, 10.6] & 0.54 & 0.45 \\
Kimi K3 & mid & medium & 5.6 [3.2, 8.4] & 29.3 [12.0, 48.0] & 11.1 [6.6, 15.8] & 1.42 & 2.09 \\
Kimi K3 & top & high & 5.4 [2.8, 8.4] & 20.0 [6.7, 36.0] & 8.8 [4.9, 13.2] & 1.49 & 1.39 \\
Gemini 3.8 Flash & low & low & 7.2 [3.2, 12.2] & 22.7 [5.3, 42.7] & 10.8 [5.7, 16.6] & 1.61 & 1.71 \\
Gemini 3.8 Flash & mid & medium & 5.2 [2.0, 9.0] & 22.7 [5.3, 42.7] & 9.2 [4.5, 14.6] & 1.31 & 1.52 \\
Gemini 3.8 Flash & top & high & 4.0 [1.0, 8.4] & 24.0 [8.0, 42.7] & 8.6 [4.0, 14.0] & 1.29 & 1.68 \\
\bottomrule
\end{tabular}
\end{adjustbox}
\caption{Complete performance metrics across evaluated lanes. Pass rates in percent with 95\% confidence intervals and cost per attempt are shown for every system and reasoning setting.}
\label{tab:all-lanes}
\end{table}

\clearpage
\section{Failure Categories and Attempt Outcomes}
\label{app:failures}

\begin{figure}[h]
\centering
\includegraphics[width=\linewidth]{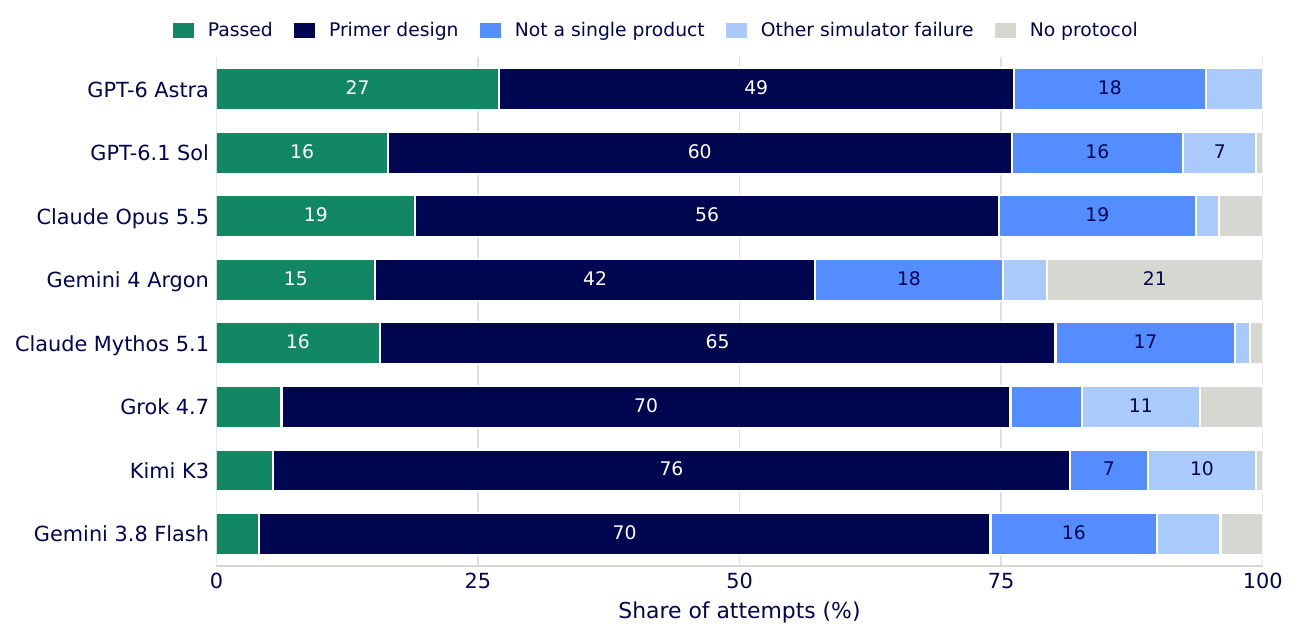}
\caption{First blocking failure of protocol-development attempts at top effort. Shares of graded attempts are shown per system, with counts in \tabref{tab:failure-categories}.}
\label{fig:failure-categories}
\end{figure}

\begin{table}[h]
\centering
\scriptsize
\begin{adjustbox}{max width=\linewidth}
\begin{tabular}{lrrrrrrrr}
\toprule
\textbf{System} & \textbf{Graded} & \textbf{Passed} & \shortstack{\textbf{Primer}\\\textbf{design}} & \shortstack{\textbf{Final tube not}\\\textbf{single product}} & \shortstack{\textbf{Bench step}\\\textbf{failed}} & \shortstack{\textbf{Wrong}\\\textbf{sequence}} & \shortstack{\textbf{Parser}\\\textbf{rejected}} & \shortstack{\textbf{No}\\\textbf{protocol}} \\
\midrule
GPT-6 Astra & 500 & 135 & 246 & 92 & 9 & 5 & 13 & 0 \\
GPT-6.1 Sol & 500 & 82 & 298 & 82 & 23 & 7 & 5 & 3 \\
Claude Opus 5.5 & 500 & 95 & 279 & 94 & 2 & 0 & 9 & 21 \\
Gemini 4 Argon & 500 & 76 & 210 & 90 & 7 & 0 & 14 & 103 \\
Claude Mythos 5.1 & 500 & 78 & 323 & 86 & 0 & 3 & 4 & 6 \\
Grok 4.7 & 498 & 31 & 347 & 34 & 23 & 0 & 33 & 30 \\
Kimi K3 & 500 & 27 & 381 & 37 & 32 & 4 & 16 & 3 \\
Gemini 3.8 Flash & 496 & 20 & 347 & 79 & 15 & 5 & 10 & 20 \\
\bottomrule
\end{tabular}
\end{adjustbox}
\caption{First blocking failures by system for protocol-development tasks at top effort.}
\label{tab:failure-categories}
\end{table}

\begin{table}[h]
\centering
\scriptsize
\begin{adjustbox}{max width=\linewidth}
\begin{tabular}{lrrrrrrr}
\toprule
\textbf{System} & \textbf{Recorded} & \textbf{Graded} & \textbf{No protocol} & \textbf{Time cap} & \textbf{Refused} & \shortstack{\textbf{Parser}\\\textbf{rejected}} & \textbf{Unscored} \\
\midrule
GPT-6 Astra & 500 & 500 & 0 & 0 & 0 & 13 & 0 \\
GPT-6.1 Sol & 500 & 500 & 3 & 13 & 0 & 5 & 0 \\
Claude Opus 5.5 & 500 & 500 & 21 & 18 & 0 & 9 & 0 \\
Gemini 4 Argon & 500 & 500 & 103 & 1 & 104 & 14 & 0 \\
Claude Mythos 5.1 & 500 & 500 & 6 & 3 & 0 & 4 & 0 \\
Grok 4.7 & 548 & 498 & 30 & 30 & 0 & 33 & 50 \\
Kimi K3 & 501 & 500 & 3 & 2 & 0 & 16 & 1 \\
Gemini 3.8 Flash & 500 & 496 & 20 & 23 & 0 & 10 & 4 \\
\bottomrule
\end{tabular}
\end{adjustbox}
\caption{Breakdown of execution and grading statuses for protocol-development attempts.}
\label{tab:attempt-outcomes}
\end{table}

\clearpage
\section{Reviewed Construct-Design Submissions}
\label{app:reviewed}

Adjudication outcomes are summarized in \tabref{tab:adjudication}, and every reviewed attempt at the top reasoning setting is listed in \tabref{tab:reviewed-attempts}. \tabref{tab:judge-gate} gives, for the ten judged tasks, how many top-effort attempts passed the deterministic checks and how many of those the judge scored below 0.9. Review decides only whether a submission passes the deterministic checks. On the ten judged tasks the final score also includes the rubric judge, so a reviewed attempt can pass the deterministic checks and still fall below the 0.9 threshold; five top-effort attempts did (four by Gemini 4 Argon and one by Gemini 3.8 Flash), and they are counted as failures in \tabrefs{tab:results}{tab:construct-grid}.

\begin{table}[h]
\centering
\small
\begin{tabularx}{\linewidth}{Yrr}
\toprule
\textbf{Adjudication outcome} & \textbf{Top-tier trials (8$\times$75)} & \textbf{All 24 lanes (24$\times$75)} \\
\midrule
Shipped grader result stands & 552 & 1626 \\
A: passes after value-preserving re-expression of the record & 17 & 43 \\
A$'$: engine source checks verify product and route as written & 5 & 7 \\
B: passes after manual verification of the engine evidence & 5 & 26 \\
0: still fails after allowed re-expression & 21 & 95 \\
Unscored (harness failure) & 0 & 3 \\
\midrule
\textbf{Total} & \textbf{600} & \textbf{1800} \\
\bottomrule
\end{tabularx}
\caption{Adjudication outcomes for construct-design submissions. Counts are shown for the eight top-effort lanes and for all 24 lanes.}
\label{tab:adjudication}
\end{table}

\begin{table}[p]
\centering
\scriptsize
\renewcommand{\arraystretch}{0.9}
\begin{tabularx}{\linewidth}{llccY}
\toprule
\textbf{System} & \textbf{Task} & \textbf{Deterministic} & \textbf{Final} & \textbf{Basis} \\
\midrule
Claude Mythos 5.1 & AG/AP1 chimera plasmid & pass & pass & passes after the record is re-expressed without changing any value \\
Claude Opus 5.5 & AG/AP1 chimera plasmid & pass & pass & passes after the record is re-expressed without changing any value \\
Gemini 4 Argon & AG/AP1 chimera plasmid & pass & pass & passes on manual check of the grader's evidence \\
Gemini 4 Argon & AG/AP1 chimera plasmid & pass & pass & passes on manual check of the grader's evidence \\
Gemini 4 Argon & AG/AP1 chimera plasmid & fail & fail & still fails after allowed re-expression \\
Gemini 4 Argon & AG/AP1 chimera plasmid & pass & pass & passes on manual check of the grader's evidence \\
Gemini 4 Argon & AG/AP1 chimera plasmid & pass & pass & passes on manual check of the grader's evidence \\
GPT-6.1 Sol & Alkaloid-pathway GoldenBraid set & pass & pass & grader's own source checks verify the product and route as written \\
Gemini 4 Argon & Alkaloid-pathway GoldenBraid set & pass & fail (0.79) & passes on manual check of the grader's evidence \\
Claude Opus 5.5 & GFP into EcoFlex vector & pass & pass & passes after the record is re-expressed without changing any value \\
Claude Opus 5.5 & GFP into EcoFlex vector & pass & pass & passes after the record is re-expressed without changing any value \\
GPT-6 Astra & GFP into EcoFlex vector & pass & pass & passes after the record is re-expressed without changing any value \\
GPT-6 Astra & GFP into EcoFlex vector & pass & pass & passes after the record is re-expressed without changing any value \\
GPT-6.1 Sol & GFP into EcoFlex vector & pass & pass & passes after the record is re-expressed without changing any value \\
Gemini 4 Argon & GFP into EcoFlex vector & pass & pass & passes after the record is re-expressed without changing any value \\
Gemini 4 Argon & GFP into EcoFlex vector & pass & pass & passes after the record is re-expressed without changing any value \\
Gemini 4 Argon & GFP into EcoFlex vector & pass & pass & passes after the record is re-expressed without changing any value \\
Grok 4.7 & GFP into EcoFlex vector & pass & pass & grader's own source checks verify the product and route as written \\
Grok 4.7 & GFP into EcoFlex vector & pass & pass & grader's own source checks verify the product and route as written \\
Grok 4.7 & GFP into EcoFlex vector & pass & pass & grader's own source checks verify the product and route as written \\
Claude Mythos 5.1 & Prime-assembly donor family & fail & fail & still fails after allowed re-expression \\
Claude Mythos 5.1 & Prime-assembly donor family & fail & fail & still fails after allowed re-expression \\
Claude Opus 5.5 & Prime-assembly donor family & fail & fail & still fails after allowed re-expression \\
Claude Opus 5.5 & Prime-assembly donor family & fail & fail & still fails after allowed re-expression \\
GPT-6.1 Sol & Prime-assembly donor family & pass & pass & passes after the record is re-expressed without changing any value \\
Gemini 3.8 Flash & Prime-assembly donor family & fail & fail & still fails after allowed re-expression \\
Gemini 3.8 Flash & Prime-assembly donor family & pass & fail (0.67) & passes after the record is re-expressed without changing any value \\
Gemini 3.8 Flash & Prime-assembly donor family & fail & fail & still fails after allowed re-expression \\
Gemini 3.8 Flash & Prime-assembly donor family & fail & fail & still fails after allowed re-expression \\
Gemini 3.8 Flash & Prime-assembly donor family & fail & fail & still fails after allowed re-expression \\
Gemini 4 Argon & Prime-assembly donor family & pass & fail (0.80) & passes after the record is re-expressed without changing any value \\
Gemini 4 Argon & Prime-assembly donor family & pass & fail (0.81) & passes after the record is re-expressed without changing any value \\
Gemini 4 Argon & Prime-assembly donor family & pass & fail (0.88) & passes after the record is re-expressed without changing any value \\
Grok 4.7 & Prime-assembly donor family & fail & fail & still fails after allowed re-expression \\
Grok 4.7 & Prime-assembly donor family & fail & fail & still fails after allowed re-expression \\
Grok 4.7 & Prime-assembly donor family & fail & fail & still fails after allowed re-expression \\
Kimi K3 & Prime-assembly donor family & fail & fail & still fails after allowed re-expression \\
Kimi K3 & Prime-assembly donor family & fail & fail & still fails after allowed re-expression \\
Kimi K3 & Prime-assembly donor family & fail & fail & still fails after allowed re-expression \\
Kimi K3 & Prime-assembly donor family & fail & fail & still fails after allowed re-expression \\
Claude Opus 5.5 & Synthetic-chromosome locus edit & pass & pass & grader's own source checks verify the product and route as written \\
Claude Mythos 5.1 & Tus/TerB HEK293 constructs & fail & fail & still fails after allowed re-expression \\
Claude Mythos 5.1 & Tus/TerB HEK293 constructs & pass & pass & passes after the record is re-expressed without changing any value \\
GPT-6.1 Sol & Tus/TerB HEK293 constructs & pass & pass & passes after the record is re-expressed without changing any value \\
Gemini 3.8 Flash & Tus/TerB HEK293 constructs & fail & fail & still fails after allowed re-expression \\
Gemini 3.8 Flash & Tus/TerB HEK293 constructs & fail & fail & still fails after allowed re-expression \\
Gemini 4 Argon & Tus/TerB HEK293 constructs & fail & fail & still fails after allowed re-expression \\
Grok 4.7 & Tus/TerB HEK293 constructs & fail & fail & still fails after allowed re-expression \\
\bottomrule
\end{tabularx}
\caption{Reviewed construct-design attempts at top effort. Deterministic results after review and final results at the 0.9 threshold are listed with the basis for each decision.}
\label{tab:reviewed-attempts}
\end{table}

\begin{table}[h]
\centering
\small
\begin{adjustbox}{max width=\linewidth}
\begin{tabular}{lrrr}
\toprule
\textbf{System} & \textbf{Attempts on judged tasks} & \textbf{Passed deterministic checks} & \textbf{Judge score below 0.9} \\
\midrule
GPT-6 Astra & 50 & 28 & 2 \\
GPT-6.1 Sol & 50 & 34 & 6 \\
Claude Opus 5.5 & 50 & 9 & 3 \\
Gemini 4 Argon & 50 & 21 & 8 \\
Claude Mythos 5.1 & 50 & 6 & 3 \\
Grok 4.7 & 50 & 18 & 3 \\
Kimi K3 & 50 & 15 & 6 \\
Gemini 3.8 Flash & 50 & 11 & 3 \\
\midrule
\textbf{Total} & \textbf{400} & \textbf{142} & \textbf{34} \\
\bottomrule
\end{tabular}
\end{adjustbox}
\caption{Judge threshold on the ten rubric-judged tasks at top effort. Attempts that passed the deterministic checks and, of those, the number the judge scored below 0.9 are shown per system.}
\label{tab:judge-gate}
\end{table}

\clearpage
\section{Repeated Sampling and Per-Task Results}
\label{app:per-task}

Confidence intervals in \tabref{tab:pass-at-k} use the same bootstrap draws as \tabref{tab:results}.

\begin{table}[h]
\centering
\scriptsize
\begin{adjustbox}{max width=\linewidth}
\begin{tabular}{lcccrrl}
\toprule
\textbf{System} & \textbf{pass@1 \% [95\% CI]} & \textbf{pass@5 \% [95\% CI]} & \textbf{pass@10 \% [95\% CI]} & \textbf{Solved $\geq$1} & \textbf{Solved all} & \textbf{Note} \\
\midrule
GPT-6 Astra & 27.0 [19.8, 34.6] & 59.9 [48.3, 71.0] & 72.0 [60.0, 84.0] & 36 & 2 & \\
GPT-6.1 Sol & 16.4 [11.4, 22.2] & 48.3 [37.7, 58.8] & 66.0 [52.0, 78.0] & 33 & 0 & \\
Claude Opus 5.5 & 19.0 [12.0, 27.0] & 44.6 [33.4, 56.0] & 58.0 [44.0, 72.0] & 29 & 3 & \\
Gemini 4 Argon & 15.2 [8.4, 23.2] & 32.4 [21.4, 44.0] & 42.0 [28.0, 56.0] & 21 & 2 & \\
Claude Mythos 5.1 & 15.6 [9.6, 22.6] & 42.5 [32.1, 52.9] & 60.0 [46.0, 74.0] & 30 & 2 & \\
Grok 4.7 & 6.2 [3.4, 9.2] & 22.9 [13.8, 32.6] & 34.0 [22.0, 48.0] & 17 & 0 & $k=9$ cap (2 tasks) \\
Kimi K3 & 5.4 [2.8, 8.4] & 20.3 [12.1, 29.4] & 32.0 [20.0, 46.0] & 16 & 0 & \\
Gemini 3.8 Flash & 4.0 [1.0, 8.4] & 10.5 [4.0, 18.1] & 16.0 [6.0, 26.0] & 8 & 0 & $k=9$ cap (4 tasks) \\
\bottomrule
\end{tabular}
\end{adjustbox}
\caption{Protocol-development \passatk{} and task coverage at top effort. Pass rates in percent with 95\% confidence intervals are shown at $k=1$, $5$, and $10$ with the number of tasks solved at least once and in every attempt.}
\label{tab:pass-at-k}
\end{table}

\begin{table}[h]
\centering
\scriptsize
\begin{adjustbox}{max width=\linewidth}
\begin{tabular}{lcccccccc}
\toprule
\textbf{Task} & \shortstack{\textbf{GPT-6}\\\textbf{Astra}} & \shortstack{\textbf{GPT-6.1}\\\textbf{Sol}} & \shortstack{\textbf{Claude}\\\textbf{Opus 5.5}} & \shortstack{\textbf{Gemini 4}\\\textbf{Argon}} & \shortstack{\textbf{Claude}\\\textbf{Mythos 5.1}} & \shortstack{\textbf{Grok}\\\textbf{4.7}} & \shortstack{\textbf{Kimi}\\\textbf{K3}} & \shortstack{\textbf{Gemini 3.8}\\\textbf{Flash}} \\
\midrule
Matched promoter panel & 0/5 & 0/5 & 5/5 & 5/5 & 4/5 & 5/5 & 3/5 & 3/5 \\
GFP into EcoFlex vector & 5/5 & 5/5 & 3/5 & 3/5 & 1/5 & 3/5 & 1/5 & 0/5 \\
Insulin intermediate construct & 5/5 & 4/5 & 0/5 & 1/5 & 0/5 & 4/5 & 2/5 & 5/5 \\
Minimal-edit recoding & 5/5 & 5/5 & 1/5 & 0/5 & 0/5 & 5/5 & 5/5 & 0/5 \\
Reporter choice from data & 3/5 & 2/5 & 5/5 & 0/5 & 3/5 & 0/5 & 3/5 & 5/5 \\
Insulin two-chain constructs & 5/5 & 5/5 & 0/5 & 5/5 & 0/5 & 2/5 & 1/5 & 2/5 \\
MPRA construct rebuild & 3/5 & 1/5 & 3/5 & 5/5 & 3/5 & 0/5 & 0/5 & 2/5 \\
AG/AP1 chimera plasmid & 2/5 & 2/5 & 1/5 & 4/5 & 1/5 & 1/5 & 0/5 & 1/5 \\
Detector choice from antibody data & 3/5 & 4/5 & 0/5 & 0/5 & 0/5 & 0/5 & 0/5 & 0/5 \\
Alkaloid-pathway GoldenBraid set & 1/5 & 1/5 & 0/5 & 0/5 & 0/5 & 0/5 & 0/5 & 0/5 \\
Tus/TerB HEK293 constructs & 0/5 & 1/5 & 0/5 & 0/5 & 1/5 & 0/5 & 0/5 & 0/5 \\
Prime-assembly donor family & 0/5 & 1/5 & 0/5 & 0/5 & 0/5 & 0/5 & 0/5 & 0/5 \\
Synthetic-chromosome locus edit & 0/5 & 0/5 & 1/5 & 0/5 & 0/5 & 0/5 & 0/5 & 0/5 \\
Sodium-channel promoter repair & 0/5 & 0/5 & 0/5 & 0/5 & 0/5 & 0/5 & 0/5 & 0/5 \\
Dual-luciferase frame panel & 0/5 & 0/5 & 0/5 & 0/5 & 0/5 & 0/5 & 0/5 & 0/5 \\
\bottomrule
\end{tabular}
\end{adjustbox}
\caption{Construct-design solves per task and system at top effort. Each cell gives the number of five attempts that passed.}
\label{tab:construct-grid}
\end{table}

\end{document}